\documentclass[letterpaper, 10 pt, conference]{ieeeconf}  % Comment this line out if you need a4paper

\IEEEoverridecommandlockouts                              % This command is only needed if 
\usepackage{amsmath}                % for align environment
\usepackage{amssymb}                % symbols such as \Box
\usepackage[dvipsnames]{xcolor}                 % for \textcolor{red}{}
\usepackage{bm}         % for proper bold vectors + bold greek letters
\usepackage{graphicx}
\usepackage{booktabs}

\usepackage{comment}

\let\vec\bm

\newtheorem{definition}{Definition}

\newtheorem{assumption}{Assumption}

\usepackage{hyperref}
\hypersetup{%
  colorlinks=true,%
  linkcolor={red!50!black},
  citecolor={blue!65!black},
  urlcolor={blue!80!black},
  bookmarksnumbered=true,%
  bookmarksopen=true}
\usepackage[capitalise]{cleveref}
\usepackage{caption}
\makeatletter
\let\oldsine\sin
\let\oldcos\cos

\renewcommand{\sin}{\@ifnextchar\bgroup\sin@arg\oldsine}
\renewcommand{\cos}{\@ifnextchar\bgroup\cos@arg\oldcos}

\newcommand{\sin@arg}[1]{s_{#1}}
\newcommand{\cos@arg}[1]{c_{#1}}
\makeatother

\newcommand{\norm}[1]{\left\lVert#1\right\rVert}

\newcommand{\config}{\boldsymbol{\eta}}
\newcommand{\twist}{\boldsymbol{\nu}}

\newcommand{\pos}{\boldsymbol{p}}
\newcommand{\quat}{\boldsymbol{\mathfrak{q}}}
\newcommand{\vel}{\boldsymbol{v}}
\newcommand{\angvel}{\boldsymbol{\omega}}

\newcommand{\wrench}{\mathbf{w}}
\newcommand{\torque}{\boldsymbol{\tau}}

\newcommand{\inertia}{\boldsymbol{M}}
\newcommand{\coriolis}{\boldsymbol{C}}

\newcommand{\gravity}{\boldsymbol{g}}

\title{\LARGE \bf
Validation of Space Robotics in Underwater Environments via Disturbance Robustness Equivalency}

\author{Joris Verhagen$^{*1}$, Elias Krantz$^{2}$, Chelsea Sidrane$^{1}$, David D\"orner$^{2}$, Nicola De Carli$^{3}$, Pedro Roque$^{3,4}$, \\ Huina Mao$^{2}$, Gunnar Tibert$^{2}$, Ivan Stenius$^{2}$, Christer Fuglesang$^{2}$, Dimos V. Dimarogonas$^{3}$, Jana Tumova$^{1}$ 
\thanks{$^{*}$Corresponding author: {\tt\small jorisv@kth.se}}
\thanks{$^{1}$Division of Robotics, Perception and Learning, KTH Royal Institute of Technology, Stockholm, Sweden.}
\thanks{$^{2}$School of Engineering Sciences, KTH Royal Institute of Technology, Stockholm, Sweden.}
\thanks{$^{3}$Division of Decision and Control Systems, KTH Royal Institute of Technology, Stockholm, Sweden.}
\thanks{$^{4}$ Department of Mechanical and Civil Engineering, California Institute of Technology, Pasadena, United States.
\newline{
This work was partially supported by the Wallenberg AI, Autonomous Systems and
Software Program (WASP) funded by the Knut and Alice Wallenberg Foundation and by the Swedish Research Council (VR) grant number 2024-05701. The authors are also affiliated with Digital Futures.
}}
}

\begin{document}

\maketitle
% \IEEEpeerreviewmaketitle

\thispagestyle{empty}
\pagestyle{empty}

%%%%%%%%%%%%%%%%%%%%%%%%%%%%%%%%%%%%%%%%%%%%%%%%%%%%%%%%%%%%%%%%%%%%%%%%%%%%%%%%
\begin{abstract}

%We present a complete, rigorous approach to the validation of space robots using underwater environments. 
%We start from the perspective where a space platform has been given a high-level specification to satisfy and we wish to verify whether the space platforms robustness is sufficient to carry out this task while only being able to control an underwater equivalent platform. 
%We consider the different dynamics of both systems and environments to synthesize a specification that is equivalent in difficulty for both systems. 
%\ND{I would emphasize a lot on the experimental validation. Below is my proposition:
%}

%{\color{magenta}
We present an experimental validation framework for space robotics that leverages underwater environments to approximate microgravity dynamics. 
While neutral buoyancy conditions make underwater robotics an excellent platform for space robotics validation, there are still dynamical and environmental differences that need to be overcome. 
Given a high-level space mission specification, expressed in terms of a Signal Temporal Logic specification, we overcome these differences via the notion of \emph{maximal disturbance robustness} of the mission. We formulate the motion planning problem such that the original space mission and the validation mission achieve the same disturbance robustness degree.
The validation platform then executes its mission plan using a near-identical control strategy to the space mission where the closed-loop controller considers the spacecraft dynamics.
Evaluating our validation framework relies on estimating disturbances during execution and comparing them to the disturbance robustness degree, providing practical evidence of operation in the space environment.
Our evaluation features a dual-experiment setup: an underwater robot operating under near-neutral buoyancy conditions to validate the planning and control strategy of either an experimental planar spacecraft platform or a CubeSat in a high-fidelity space dynamics simulator.

Code and videos can be found on the \href{https://DISCOWER.github.io/space_validation}{project page}
% Our evaluation features a dual-experiment setup: a free-flying spacecraft analog platform
% ,equipped with air bearings and thrusters to simulate two-dimensional motion control, 
% and an underwater robot operating under near-neutral buoyancy conditions to extend this simulation to three-dimensional spacecraft motion control. 

% Multiple mission objectives are tested on each robot. The objectives are expressed in terms of Signal Temporal Logic (STL) specifications, ensuring that both robots follow trajectories satisfying the same formal requirements. To enable fair comparison, we introduce a notion of \emph{maximal disturbance robustness} of the mission and formulate the motion planning problem so that the underwater mission is adjusted to achieve the same disturbance robustness degree as the space mission. Both robots execute the their mission plan under 
% %(ALMOST, DUE TO DYNAMICS CANCELLATION) 
% near-identical control strategies, with actuator limits matched to ensure comparable task difficulty. This validation highlights the extent to which underwater experiments can capture the challenges of space robotics, and vice versa, providing practical evidence of their suitability for testing autonomy, control, and mission execution in microgravity-like conditions.%}
% % Using model-based robustness metrics in spatio-temporal planning allows us to synthesise and map plans to ensure the space and underwater platform have a specification of equal difficulty.

\end{abstract}

%%%%%%%%%%%%%%%%%%%%%%%%%%%%%%%%%%%%%%%%%%%%%%%%%%%%%%%%%%%%%%%%%%%%%%%%%%%%%%%%

\section{INTRODUCTION}
The significant costs and safety-critical nature of autonomous space missions make verification and validation essential steps in the development pipeline. However, faithfully reproducing spacecraft operation conditions on Earth remains a longstanding challenge. Several terrestrial validation platforms have been developed to address this, including granite tables \cite{uyama2010integrated} or epoxy resin floors \cite{roque2025towards} serving as calibrated flat surfaces where robotic systems are supported on air bearings to approximate undamped planar motion. 
Higher degrees-of-freedom (DoF) weightlessness has been emulated using robotic arms with specialized joints~\cite{saulnier2014six,tsiotras2014astros}.
% For approximating higher degrees-of-freedom (DoF) weightlessness, additional robotic arms mounted on specialized joints have been used~\cite{saulnier2014six,tsiotras2014astros}. 
While these facilities provide valuable insights, their accuracy for testing true 6-DoF motion is limited due to the difficulty of emulating weightlessness.
Moreover, such platforms are impractical for evaluating human–robot collaboration tasks, which increasingly play a role in future space missions.

As an alternative, underwater environments have emerged as a promising surrogate for space. Neutral buoyancy experiments are already a cornerstone of astronaut training, with large-scale facilities such as NASA’s Neutral Buoyancy Laboratory~\cite{jairala2012eva} and the Neutral Buoyancy Research Facility~\cite{akin_robotic_capabilities} (NBRF) at the University of Maryland supporting both human operations and autonomous system testing. The appeal of underwater environments lies in their ability to approximate aspects of microgravity: a vehicle close to neutral buoyancy experiences significantly reduced weight effects, and the surrounding fluid environment induces dynamics with important parallels to those of free-flying robots in space. These commonalities have motivated the use of underwater systems for prototyping, training, and system-level evaluation of space technologies.
% such as at the \href{https://robotics.umd.edu/facilities/neutral-buoyancy-research-facility}{Neutral Buoyancy lab} at the University of Maryland.

\begin{figure}[t!]
    \centering
    \includegraphics[width=0.38\textwidth]{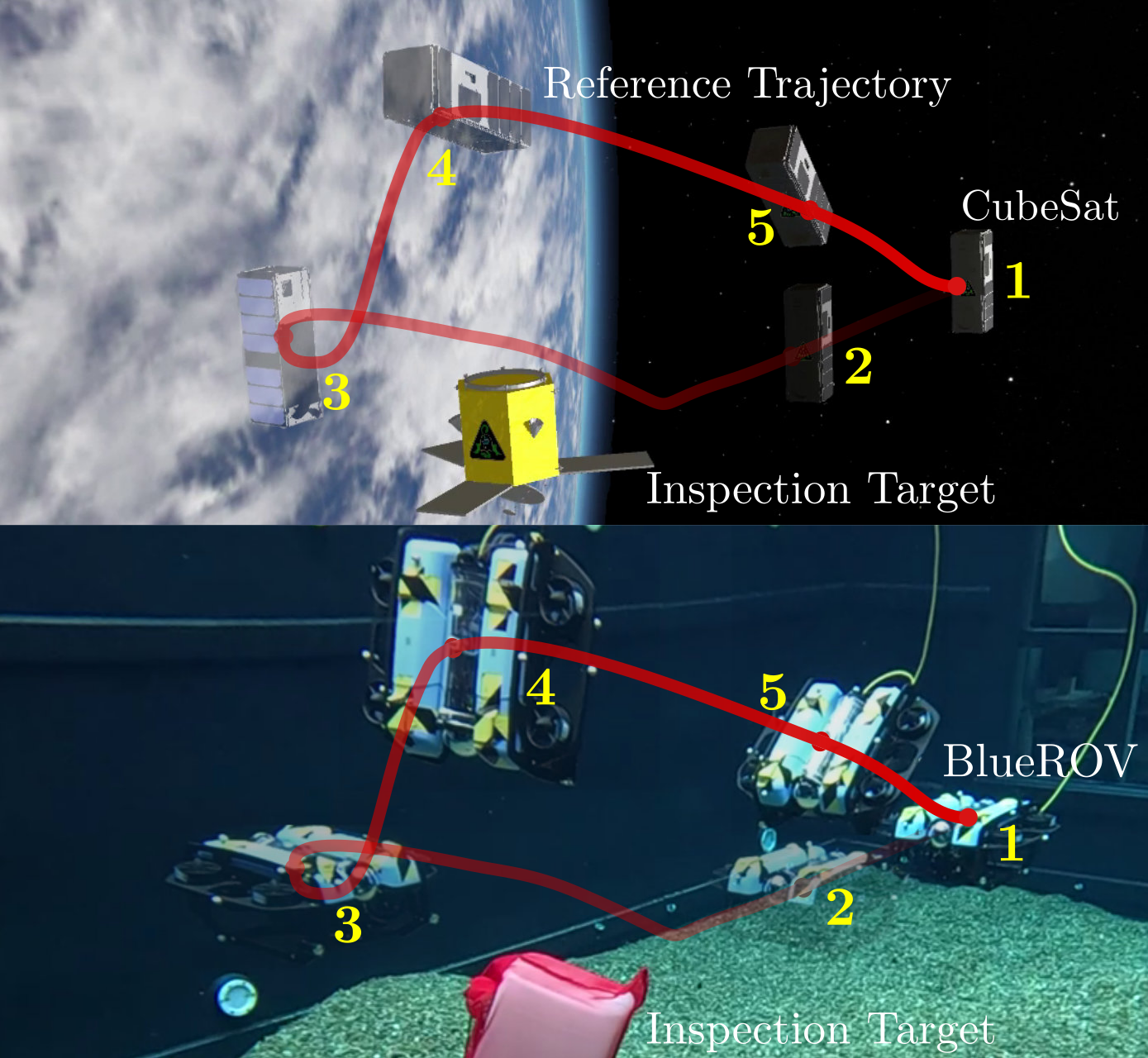}
     \vspace{-0.1cm}
    \caption{A physical BlueROV underwater robot is utilized to validate a planner and controller which aim to inspect a passive target in space. While the environments are distinct, they share sufficient similarities in order to be used as a validation platform.}
    \label{fig:intro_figure}
    \vspace{-0.6cm}
\end{figure}

Several underwater spacecraft analogs have been developed toward this goal. At the Space Systems Laboratory (SSL), free-flying underwater vehicles such as SCAMP~\cite{miller_attitude_2000}, EUCLID~\cite{nasa_vertical_2014}, and the Ranger Neutral Buoyancy Vehicle (NBV)~\cite{carignan_reaction_stabilization} were deployed in NBRF. SCAMP, finalized in 1997, was primarily teleoperated and used for inspection, extra-vehicular activity support, and situational awareness, while later work incorporated vision-based closed-loop control for semi-autonomous station keeping and position regulation. Ranger NBV and EUCLID extended this line of research toward servicing and multi-vehicle operations.
% More recently, commercially available platforms such as the BlueROV~\cite{von2022open} have broadened access to underwater robotics research, offering a modular and open-source basis that complements the earlier SSL testbeds.
% Beyond SSL, NASA’s NEEMO program at the Aquarius habitat and the Pavilion Lake Research Project explored underwater analogs from an operational perspective, using AUVs (Autonomous Underwater Vehicles), submersibles, and divers to evaluate exploration architectures and human–robot coordination in simulated space mission scenarios~\cite{trembanis2012multiplatform,forrest2010performance}.

Despite their similarities, key challenges remain in making underwater experiments a systematic validation tool for space robotics. Most notably, the differences in dynamics between the two environments during mission execution introduce uncertainties about how performance in water translates to performance in space. 
To bridge this gap, we argue that validation requires adapting the mission specification in a principled way to produce \textit{equivalent} behavior across robots. 
In particular, when a mission is specified in terms of formal requirements -- such as those expressed in Signal Temporal Logic (STL)~\cite{donze2010robust} -- the generated trajectories must not only satisfy these requirements but also exhibit a comparable degree of robustness with respect to disturbances. 

A comparable line of research addresses cross-platform validation via formal abstractions such as simulation relations~\cite{tabuada2009verification}, relative reachability~\cite{herbert2017fastrack} or maneuver automata~\cite{hess2014formal}. This establishes an equivalence between systems at the level of discrete behaviors. 
Our goal instead is to, with minimal modification to the planning and control stack of the space platform, validate its operation in a different environment without conservativeness and with global optimality on complex task specifications.
Our approach subsequently relies on equivalence through a planner (via disturbance robustness degrees) and a closed-loop controller (via feedback equivalence).

\subsection*{Contributions}
The central idea of this work is that underwater testing can be used to validate spacecraft motion planners and controllers for on-orbit tasks. Our specific contributions are:

\begin{itemize}
    % \item \textbf{Cross-environment evaluation framework.} 
    % We propose an evaluation framework where space mission planners and controllers can be evaluated on a validation platform (in this case an underwater robot). 
    % We execute the same STL-based mission specifications underwater and on two reference platforms to evaluate the underwater results: a planar spacecraft analog (hereafter referred to as the \emph{space robot}, designed to experimentally mimic spacecraft 2D-dynamics) and the Basilisk astrodynamics simulation framework~\cite{kenneally2020basilisk}. 
    % All platforms use Model Predictive Control (MPC) with an offset-free formulation~\cite{pannocchia_disturbance_2003} to ensure accurate steady-state tracking.
    \item \textbf{Robustness-matched planning.} We extend the notion of \emph{disturbance robustness}~\cite{verhagen2024robust} to capture the maximal disturbances that can be counteracted during the mission. We adjust the mission specification so that its robustness in underwater settings matches the robustness associated with execution in space. Thereby, we ensure comparability of plans across the two environments.

    % \item \textbf{Feedback-equivalent control.} We are able to validate a control architecture on the underwater robot, which is nearly identical to its space counterpart via feedback equivalence and Model Predictive Control (MPC), which supports consistent execution of the same planned trajectories across environments. We additionally estimate disturbances acting on the underwater robot and assess whether they fall within the disturbance robustness degree from the planner.

    \item \textbf{Feedback-equivalent control.} We validate on the underwater robot a control architecture that is nearly identical to its space counterpart through feedback equivalence and Model Predictive Control (MPC), enabling consistent execution of the same planned trajectories across environments. We additionally estimate disturbances acting on the underwater robot and assess whether they remain within the disturbance robustness degree from the planner, thereby ensuring that underwater results are representative of the intended space performance.

    % \item \textbf{Feedback-equivalent control.} We realize near-identical control architectures via feedback equivalence and Model Predictive Control (MPC), which supports consistent execution of the same planned trajectories across environments.

    % \item \textbf{Disturbance assessment.} During underwater execution, we estimate the disturbances acting on the robot and assess whether they fall within the disturbance robustness degree obtained from the planner.

    \item \textbf{Experimental validation.} We evaluate the validation framework underwater using a BlueROV and two space robot platforms: a physical planar spacecraft analog and a CubeSat in an astrodynamics simulator. Two region-reaching scenarios are evaluated: (i) a 2D inspection-style scenario, comparing the underwater robot with the space robot; (ii) a 3D inspection-style scenario, comparing the underwater robot with a CubeSat simulation (seen in \cref{fig:intro_figure}.) 
\end{itemize}

\section{PRELIMINARIES}
Let $\mathbb{R}$ denote the set of real numbers, $\mathbb{S}^n$ the $n$-dimensional unit sphere, $\mathbb{N}$ the set of natural numbers (including zero), and $\mathbb{B} := \{\top, \bot\}$ the set of binary values, representing \emph{true} and \emph{false}, respectively.
%Let $\mathbb{R}$, $\mathbb{S}^{n}$, $\mathbb{N}$, and $\mathbb{B}$ be the set of real, the $n$-th sphere, natural (including zero), and binary numbers, respectively, 
%where $\mathbb{B}:=\{\top,\bot\}$ denotes \emph{true} and \emph{false}. 
$\mathbb{R}_{\geq 0}$ denotes the set of nonnegative real numbers. We use $\underline{\vec{x}}$ and $\overline{\vec{x}}$ to denote the lower-bound and upper-bound that vector $\vec{x} \in \mathbb{R}^n$ may take.
Further, let $\hat{\vec{x}}$ denote the estimation of $\vec{x}$.

\subsection{Signal Temporal Logic}
\label{ssec:preliminaries_stl}

STL is a formal specification language widely used to express temporal properties of dynamical systems. An STL formula can be interpreted either qualitatively (the signal satisfies or violates the specification) or quantitatively (the extent to which the signal satisfies or violates it).

Time-bounded STL over $n$-dimensional continuous-time signals $x:\mathbb{R}_{\ge 0} \rightarrow X \subseteq \mathbb{R}^n$ is defined as follows.

\begin{definition}[Time-bounded STL]
\label{def:stl}
    Let $I=[t_1,t_2]$ be a closed bounded time interval, with $t_1,t_2 \in \mathbb{R}_{\ge 0}$ and $t_1 \leq t_2$.
    Let $\mu:X \rightarrow \mathbb{R}$ be a real-valued function, and define the Boolean predicate $p$ as
    % {\color{magenta} bold x?}
    \begin{equation}
        p = \begin{cases}
        \begin{array}{lcr}
             \top, & \mu(x)\geq 0 \\
             \bot, &  \mu(x) < 0
        \end{array}
        \end{cases}.
    \end{equation}
    STL formulas are then recursively constructed as
    \begin{equation}
    \phi := \top \mid p \mid \neg \phi \mid \phi_1 \land \phi_2 \mid \phi_1 \hspace{0.5mm} \mathcal{U}_I  \phi_2,
    \end{equation}
    where $\neg$ and $\land$ denote Boolean negation and conjunction, and $\mathcal{U}_I$ is the time-bounded Until operator.
    \end{definition}

Intuitively, $\phi_1 \mathcal{U}_I \phi_2$ requires $\phi_1$ to hold until, within interval $I$, $\phi_2$ becomes true. From this, additional temporal operators can be derived, such as the Eventually operator $\diamondsuit_I \phi = \top  \mathcal{U}_I  \phi$ (stating that $\phi$ eventually holds within $I$), and the Always operator $\Box_I \phi = \neg  \diamondsuit_I  \neg \phi$ (stating that $\phi$ holds at all times within $I$).

Beyond Boolean satisfaction, STL admits a quantitative semantics known as \emph{spatial robustness}, which evaluates to what \emph{degree} a signal satisfies or violates a specification at time $t$. Spatial robustness is defined recursively as:
\begin{subequations}
\label{eq:stl_space_robustness}
\begin{align}
    \rho_p(t,x) &= \mu(x(t)), \\
    \rho_{\neg \phi}(t,x) &= -\rho_{\phi}(t,x), \\
    \rho_{\phi_1 \land \phi_2}(t,x) &= \min(\rho_{\phi_1}(t,x),\rho_{\phi_2}(t,x)), \\
    \rho_{\phi_1 \mathcal{U}_I \phi_2}(t,x) &= \max_{\tau \in t+I}\big(\min(\rho_{\phi_2}(\tau,x),\min_{s\in[t,\tau]}\rho_{\phi_1}(s,x) ) \big).
\end{align}
\end{subequations}
The reader is referred to~\cite{donze2010robust, lindemann2025formal} for a detailed treatment of STL semantics.

% \subsection{Feedback Linearization}
\section{PROBLEM STATEMENT}
Consider the dynamics of a robot operating in space and underwater, denoted with subscripts ${sp}$ and ${uw}$, respectively. We denote the dynamics with $\circ \in \{sp,uw\}$ as,
\begin{align}\label{eq:problem_statement}
    \mathcal{M}_\circ(\vec{x}_\circ,\vec{u}_\circ) &=  \dot{\vec{x}}_\circ = f_\circ(\vec{x}_\circ)+g_\circ(\vec{x}_\circ)\vec{u}_\circ + C_\circ(\vec{x}_\circ)\vec{d}_\circ, 
    % \mathcal{M}_{uw}(\vec{x}_{uw},\vec{u}_{uw}) &= 
    % \dot{\vec{x}}_{uw} = f_{uw}(\vec{x}_{uw})+g_{uw}(\vec{x}_{uw})\vec{u}_{uw} + C_{uw}(\vec{x}_{uw})\vec{d}_{uw},
\end{align}
where we model the full 6-DoF dynamics with actuation in force and torque. 
The state is given by $\vec{x}_\circ \in \mathcal{X}_\circ \subseteq \mathbb{R}^{9} \times \mathbb{S}^3$, the control input by $\vec{u}_\circ \in \mathcal{U}_\circ \subseteq \mathbb{R}^6$, and the disturbance terms by $\vec{d}_\circ \in \mathcal{D}_\circ \subseteq \mathbb{R}^6$ which represent bounded additive uncertainties and external perturbations. 
$f_\circ: \mathcal{X}_\circ \rightarrow\mathcal{X}_\circ$ represents the drift term, $g_\circ: \mathcal{X}_\circ \rightarrow \mathbb{R}^{12\times 6}$ the the actuation matrix and $C_\circ: \mathcal{X}_\circ \rightarrow \mathbb{R}^{12 \times 6}$ the disturbance mapping.
We use $\mathcal{M}_{\circ}(\vec{x}_\circ,\vec{u}_\circ)$ as shorthand for the dynamics of a platform.
Although underwater platforms can approximate space-like conditions through near-neutral buoyancy, the two domains differ significantly in their dynamics, parameter uncertainties, and disturbance profiles. We require that satisfaction of a specification in the underwater robot, $\phi_{uw}$, under disturbances from set $\mathcal{D}_{uw}$ implies satisfaction of the space mission $\phi_{sp}$ under disturbances from set $\mathcal{D}_{sp}$.
In other words,
% To address this discrepancy, 
% %we formulate the problem as one of implied operation under observed uncertainty. Specifically, 
% we require that satisfaction of a specification $\phi$ in the underwater experiment under its maximal disturbance set $\mathcal{D}_{uw}$ implies satisfaction of the same specification in space,
% % {\color{magenta} It is not very clear what the term "observed uncertainty" means.}
\begin{equation}
    \hat{\vec{x}}_{uw} \models \phi_{uw}, \forall \vec{d}_{uw} \in \mathcal{D}_{uw} \implies \hat{\vec{x}}_{sp} \models \phi_{sp}, \forall \vec{d}_{sp} \in \mathcal{D}_{sp},
\end{equation}
where $\vec{x}_{\circ} \models \phi_{\circ}$ denotes that the trajectory $\vec{x}_{\circ}$ satisfies  STL specification $\phi_{\circ}$.
We tackle this challenge by introducing the \emph{disturbance-robustness degree} for a mission, defined as the maximal scaling factor of the disturbance set that may affect the robot while still satisfying $\phi_{\circ}$. 
Motion planning for $\vec{x}_{sp}$ is first performed to satisfy $\phi_{sp}$, and the disturbance-robustness degree is quantified. 
The specification is then transformed into an underwater counterpart $\phi_{uw}$ such that executing $\phi_{uw}$ has an equivalent disturbance-robustness degree for the underwater robot. In this way, the two environments can be compared fairly, and the underwater execution provides systematic validation of the space mission.

\section{Models}

We consider three dynamics models: a linear space robot model $\mathcal{M}_{spL}$, a nonlinear space robot model $\mathcal{M}_{sp}$, and a nonlinear underwater robot model $\mathcal{M}_{uw}$.
The nonlinear models are modeled as a rigid body with state vector \(\vec{x}_{\circ} = [\pos,\, \quat,\, \vel,\, \angvel]\), where $\circ \in \{sp, uw\}$. Here, $\pos \in \mathbb{R}^3$ is the robot position in the inertial frame, 
$\quat \in \mathbb{S}^3$ is the body attitude represented as a quaternion relative to the inertial frame, 
and $\vel, \angvel \in \mathbb{R}^3$ are the linear and angular velocities in the body frame, respectively.
The robots are actuated through body-frame force and torque inputs 
$\wrench = [\vec{F}_{\text{act}},\, \torque_{\text{act}}]$, 
generated by the robot's thrusters, each applying a force $\mu_i \in [\mu_{\min}, \mu_{\max}]$ along its axis. 
The total wrench is 
$\wrench = \boldsymbol{G} \boldsymbol{\mu}$, 
with $\boldsymbol{\mu} = [\mu_1 \;\hdots\; \mu_m]^\mathsf{T} \in \mathbb{R}^m$ being the vector of thruster forces, and 
$\boldsymbol{G} = \begin{bmatrix} \boldsymbol{G}_F^\mathsf{T} & \boldsymbol{G}_\tau^\mathsf{T} \end{bmatrix}^\mathsf{T}$ 
the control allocation matrix mapping them to body-frame wrenches.
%
% The mapping from thruster forces to the resulting body-frame wrench depends on the thruster configuration and lever arms.
The robots are also subject to external disturbances,
$\vec{d} = [\vec{F}_d,\, \torque_d]$, where $\vec{F}_d$ and $\torque_d$ are in inertial and body-frame, respectively. We define the robot configuration $\vec{\eta} = \left[\pos,\quat\right]^T$ and body-frame velocity twist $\vec{\nu} = \left[\vel, \angvel\right]^T$. The full nonlinear model of both the space and underwater robots can then be expressed in the standard form
\begin{equation}
    \begin{aligned}
        \dot{\vec{\eta}} &= \boldsymbol{J}(\vec{\eta})\vec{\nu},\\
        \boldsymbol{M}\dot{\vec{\nu}} + \boldsymbol{C}(\vec{\nu})&\vec{\nu} + \boldsymbol{D}(\vec{\nu})\vec{\nu} + \vec{g}(\vec{\eta}) = \wrench + \boldsymbol{F}(\config)\vec{d},
    \end{aligned}
    \label{eq:general_eom}
\end{equation}
where 
\begin{equation}
    \boldsymbol{J}(\config) = \begin{bmatrix}
    \boldsymbol{R}(\quat) & \boldsymbol{0}_{3\times 3} \\ \boldsymbol{0}_{3\times 3} & \frac{1}{2}\boldsymbol{E}(\quat)
\end{bmatrix},
\boldsymbol{F}(\config) = \begin{bmatrix}
            \boldsymbol{R}(\quat)^\mathsf{T} & \boldsymbol{0}_{3\times 3}
            \\
            \boldsymbol{0}_{3\times 3}
            & \boldsymbol{I}_3
        \end{bmatrix},
\end{equation}
with $\boldsymbol{J}(\config)$ mapping body-frame velocities to inertial-frame rates, with $\boldsymbol{E}(\quat)$ being a standard matrix for quaternion kinematics, and $\boldsymbol{F}(\config)$ mapping external disturbances to body-frame rates.
% $\vec{M}$ is the generalized mass–inertia matrix, $\vec{C}(\vec{\nu})$ captures Coriolis and centripetal terms, $\vec{D}(\vec{\nu})$ represents dissipative effects (e.g., drag in the underwater case, absent in space), $\vec{g}(\vec{\eta})$ accounts for restoring forces such as gravity and buoyancy, and $\boldsymbol{R}(\quat)$ is the rotation matrix.
$\vec{M}$ is the generalized mass–inertia matrix; 
$\vec{C}(\vec{\nu})$ the Coriolis and centripetal terms; 
$\vec{D}(\vec{\nu})$ the hydrodynamic damping; 
%$\vec{D}(\vec{\nu})$ the dissipative effects (e.g., drag underwater, absent in space); 
$\vec{g}(\vec{\eta})$ the restoring forces such as gravity and buoyancy; 
and $\boldsymbol{R}(\quat)$ the rotation matrix.
For detailed expressions of these terms for $\mathcal{M}_{uw}$, we refer the reader to~\cite{fossen.2021}.

To align with the model description in \eqref{eq:problem_statement}, we can express the dynamics functions for the full nonlinear model of the two robots as below
\begin{equation}
\begin{aligned}
    f_\circ(\boldsymbol{x}_\circ) &= \begin{bmatrix}
        \boldsymbol{J}(\config)
        \\
        \inertia^{-1}\left(-\coriolis(\twist)\twist - \vec{D}(\twist)\twist - \gravity(\config)\right)
    \end{bmatrix},
    \\
    g_\circ(\boldsymbol{x}_\circ) &= \begin{bmatrix}
        \boldsymbol{0}_{7\times 6}
        \\
        \inertia^{-1}
    \end{bmatrix},\quad
    C_\circ(\boldsymbol{x}_\circ) =
    \begin{bmatrix}
        \boldsymbol{0}_{7\times 6}
        \\
        \inertia^{-1} \boldsymbol{F}(\config)
    \end{bmatrix}
    .
    \end{aligned}
\end{equation}

The $\mathcal{M}_{sp}$--model reduces to a classical rigid-body model with only the inertia and Coriolis terms. In contrast, the $\mathcal{M}_{uw}$--model also includes added mass terms and the corresponding hydrodynamic coupling in the inertia and Coriolis matrices, as well as dissipative drag and restoring forces from gravity and buoyancy. 
Finally, the $\mathcal{M}_{spL}$--model, used for high-level planning, is a simplified linearized version of $\mathcal{M}_{sp}$, expressed with Euler angles and assuming decoupled translational and rotational dynamics.

\section{PLANNING}
\label{sec:planning}
We consider an STL specification that prescribes the space mission via desired behavior over state and time. 
We enable underwater validation of the space mission by generating motion plans for both the space and underwater robots that satisfy the specification with an equivalent disturbance-robustness degree. 
This approach reconciles the fundamental differences in their dynamics.

At the high-level planning, we consider a linear model of the space robot that enables globally optimal solutions w.r.t. the STL specification. 
We then reason about how this solution is sound when executed on the true nonlinear space robot system and can be locally adapted to the nonlinear underwater robot system.
% this preserves satisfaction of the specification for the true nonlinear space robot dynamics.
% When adapting the trajectory for the underwater robot, we use its nonlinear model. 
We assume that all predicate functions in $\phi_{sp}$ are piecewise-linear functions of the position and orientation.

\subsection{Disturbance Robustness as a Metric}
\label{ssec:dod}
% In order to reliably and soundly validate space planning and control algorithms in underwater environments, 
We utilize the size of the maximal disturbance set (via scalar multiplication) as an optimization metric in the planner to promote robustness against unforeseen disturbances.
The maximal disturbance set is then the largest set from which disturbances may be drawn that the platform can counteract while satisfying the specification $\phi_{\circ}$.
% while still satisfying the STL formula under worst-case disturbances from this set. \ND{Here, or before, a formal definition of "permissible disturbance set" should be given.}
%We denote the following assumptions that enable us to solve this planning problem to global optimality with Mixed-Integer Convex Programming.

%\begin{assumption}
%\label{ass:observability}
%    The systems $\mathcal{M}_{sp}$ and $\mathcal{M}_{uw}$ have full state and disturbance observability, meaning that the effect of parameter uncertainty or external disturbances can directly be observed. 
%\end{assumption}
%\JV{On deployment, this is under the assumption that the offset-free MPC converges instantaneously to the correct solution. Can we relax this assumption such as with bounded time convergence, bounded errors on estimation, etc? During deployment we could use the covariance estimate of the disturbance vector.}

At the planning stage, we assume that the controller can track the planned trajectory with reasonably small error despite external disturbances. To ensure this, we impose the following assumptions:

\begin{assumption}[Matched disturbances]
\label{ass:column_space}
The disturbance is \emph{matched}~\cite{krstic1995nonlinear}, meaning it acts through the same channels as the control input and there exists some $\boldsymbol{K}$ such that for any $d$ we can write $g(\vec{x})\boldsymbol{K}d = C(\vec{x})d$, or equivalently:
\begin{equation}
    \forall \vec{x}_\circ \in \mathcal{X}_\circ, \;\; \exists \boldsymbol{K} 
    \;\; \text{s.t.} \;\; g_\circ(\vec{x}_\circ) = C_\circ(\boldsymbol{x}_\circ)\boldsymbol{K}^{-1} .
\end{equation}
\end{assumption}

\begin{assumption}[Sufficient control authority]
\label{ass:supset}
The available control authority is large enough to counteract any worst-case disturbances:
\begin{equation}
    -\mathcal{U}_\circ \supseteq %C_\circ(\boldsymbol{x}_\circ)
    \boldsymbol{K}\mathcal{D}_\circ.
\end{equation}
% That is, if the disturbances were known, they could in principle be fully canceled by admissible control inputs.
\end{assumption}

%Without Assumptions~\ref{ass:column_space} and~\ref{ass:supset}, the planner would need to compute the set of reachable states as disturbances may not be directly counteracted.
%Computing the set of reachable states for a general disturbed linear system requires algorithmic approaches using zonotopes and polytopes~\cite{wetzlinger2025backward} or Hamilton-Jacobi-Isaacs reachability~\cite{bansal2017hamilton}. 
%These methods are not suitable for single-shot optimization planning as they require solving separate optimization problems for each candidate solution in the planning problem.
%when the DoD is of concern (\textcolor{red}{explain reason}).

From Assumptions~\ref{ass:column_space}--\ref{ass:supset}, disturbances effectively reduce the usable control authority of the robot. 
Intuitively, part of the control input must always be ``reserved" to counteract disturbances, so only a smaller portion of the nominal control set can be used for planning. We capture this via the \emph{effective} (or \emph{robust}) control set
\begin{equation}
\label{eq:effective_control}
    \overline{\mathcal{U}}_\circ = \mathcal{U}_\circ \ominus \boldsymbol{K}\mathcal{D}_\circ,
\end{equation}
where $\ominus$ denotes the Minkowski difference. This set represents all control inputs that remain available after accounting for the worst-case disturbances.

The maximal permissible disturbance set (equivalently, the minimal effective control set) that still allows the robot to satisfy its mission can be found by, following the notion of \emph{maximal disturbance robustness}~\cite{verhagen2024robust}, finding the maximal scalar disturbance robustness degree $\alpha^*$.
% To quantify the maximal permissible disturbance subject to the mission we find the maximal disturbance scalar factor (leading to a minimal effective control set) subject to $\phi_{\circ}$.
% Following the notion of \emph{maximal disturbance robustness}~\cite{verhagen2024robust}, we introduce a scalar robustness degree $\alpha^*$.

%We are interested in the minimal subset of $\bar{\mathcal{U}}_i$ that is required to satisfy a specification $\phi$. 
%This can be quantified in different ways where here we opt for the notion of maximal disturbance robustness~\cite{verhagen2024robust}. 
%For the sake of simplicity and closed-form solutions in the planner, we assume that $\mathcal{U}_i$ and $\mathcal{D}_i$ are hyperrectangular and include $\vec{0}$. We can then obtain the disturbance-robustness degree $\alpha^*$ as 
% \begin{equation}
%     DoD = (1-\alpha), \quad u_i^k \in \alpha \bar{\mathcal{U}}_i,
% \end{equation}
% subject to dynamics and mission constraints.
% For simplicity and closed-form tractability, we assume that $\mathcal{U}_i$ and $\mathcal{D}_i$ are a zero-centered polytope and hyperrectangle respectively.
% The disturbance-robustness degree can then be computed as
\begin{equation}
    \alpha^*=\max{\alpha}, \hspace{2mm} \textrm{s.t.} \quad \vec{u} \in \mathcal{U}_\circ \ominus \alpha \boldsymbol{K} \mathcal{D}_\circ, \;\; \vec{x}\models \phi_{\circ}
\end{equation}
%with the subsequent maximal disturbance set $\mathcal{D}^* = \alpha^*\mathcal{D}$.
with the maximal admissible disturbance set $\mathcal{D}^* = \alpha^* \mathcal{D}$.

This metric (i) identifies the largest disturbance set that preserves feasibility and (ii) enables transfer of the specification to a different platform—such as an underwater platform—by ensuring an equivalent \emph{degree of difficulty}. 
If online estimates of disturbances remain within \(\mathcal{D}^*_{uw}\), successful operation in the underwater domain provides evidence of feasibility in the space scenario under \(\mathcal{D}^*_{sp}\) (see Sec.~\ref{ssec:underwater_plan}).

\subsection{Space Plan Synthesis}
First, we consider  $\mathcal{M}_{spL}$ with rotational dynamics in Euler angles $\boldsymbol{\theta}$, with $\ddot{\boldsymbol{\theta}} = \boldsymbol{\alpha}_\omega$ as planning-level inputs, and neglecting nonlinear coupling.  
Translational dynamics are expressed in the world frame, and are hence linear.

In the nonlinear model, the feasible wrench set rotates with attitude since each thruster is fixed in the body frame. Feasibility is ensured by imposing conservative control bounds: the largest symmetric origin-centered polytope (within a chosen class) is always contained in the rotated wrench sets.
Our procedure (for forces, but analogous for torques) is as follows:
% \begin{itemize}
% \item Represent the feasible forces as a zonotope \cite{ziegler2012lectures},
% $\mathcal{F} = \{ R\left(\quat\right) \boldsymbol{G}_f \boldsymbol{\mu} | \boldsymbol{\mu}_{\min} \leq \boldsymbol{\mu} \leq \boldsymbol{\mu}_{\max} \}$,
% where $\boldsymbol{G}_f=[\boldsymbol{d}_1 \dots \boldsymbol{d}_m]$.
% \item Obtain the vertices of $\mathcal{F}$ by mapping each vertex of the input hypercube $\boldsymbol{\mu}_{\min} \leq \boldsymbol{\mu} \leq \boldsymbol{\mu}_{\max}$ and from those compute its polytope $\mathcal{H}$-representation \cite{fukuda2008exact} $\boldsymbol{H}_f \boldsymbol{\mu} \leq \boldsymbol{b}_f$ with $\boldsymbol{H}_f = \begin{bmatrix} \boldsymbol{h}_{f1} & \hdots & \boldsymbol{h}_{fm} \end{bmatrix}^\mathsf{T}$.
% \item Find the largest inscribed sphere (a simple variant of the Chebyshev center problem \cite{boyd2004convex}), whose radius can be efficiently computed as $\min_i (b_{fi}/\norm{\boldsymbol{h}_{fi}})$.
% \item Approximate this sphere with an inscribed polytope (e.g., cube, icosahedron), yielding tractable linear constraints for planning.
% \end{itemize}
% 
i) Represent the feasible forces as a zonotope,% \cite{ziegler2012lectures},
$\mathcal{F} = \{ R\left(\quat\right) \boldsymbol{G}_f \boldsymbol{\mu} | \boldsymbol{\mu}_{\min} \leq \boldsymbol{\mu} \leq \boldsymbol{\mu}_{\max} \}$,
where $\boldsymbol{G}_f=[\boldsymbol{d}_1 \dots \boldsymbol{d}_m]$;
ii) Obtain the vertices of $\mathcal{F}$ by mapping each vertex of the input hypercube $\boldsymbol{\mu}_{\min} \leq \boldsymbol{\mu} \leq \boldsymbol{\mu}_{\max}$ and from those compute its polytope $\mathcal{H}$-representation \cite{fukuda2008exact} $\boldsymbol{H}_f \boldsymbol{\mu} \leq \boldsymbol{b}_f$ with $\boldsymbol{H}_f = \begin{bmatrix} \boldsymbol{h}_{f1} & \hdots & \boldsymbol{h}_{fm} \end{bmatrix}^\mathsf{T}$;
iii) Find the largest inscribed sphere (a simple variant of the Chebyshev center problem \cite{boyd2004convex}), whose radius can be efficiently computed as $\min_i (b_{fi}/\norm{\boldsymbol{h}_{fi}})$;
iv) Approximate this sphere with an inscribed polytope (e.g., cube, icosahedron), yielding tractable linear constraints for planning.

% \ND{TODO: Maybe add plot from python}
In what follows, $\mathcal{U}_{spL}$ is such an inscribed polytope.
This linearized formulation enables us to cast the planning problem—maximizing disturbance robustness under STL specifications for the space robot—as a Mixed-Integer Linear Program.
\begin{subequations}
\label{eq:problem-def}
\begin{align}
        \min_{\vec{x}_{spL}, \vec{u}_{spL}, \alpha} & -\alpha - c_1\rho_{\phi_{sp}}(t,\vec{x}_{spL}) + c_2J(\vec{u}_{spL}), \label{obj}\\
        \text{s.t.} \quad & \vec{x}_{spL}^{k+1} = \vec{x}_{spL}^{k} + \mathcal{M}_{spL}(\vec{x}_{spL}^{k},\vec{u}_{spL}^k)\Delta t,  \label{dyn}\\
        \quad & \vec{u}_{spL} \in \mathcal{U}_{spL} \ominus \alpha \boldsymbol{K} \mathcal{D}_{spL}, \label{con}
\end{align}
\end{subequations}
% \ND{It is good to maximize robustness, but like this, with this multi-objective optimization, it is possible that we do not achieve task satisfaction even if it would be feasible. Should we add a constraint $\rho(x)>0$?}
% \ND{If it is forward Euler then it's better to write explicitly the expression rather than introducing a new symbol}
% where $\text{Euler}$ denotes an Euler integration scheme with a $\Delta t$ time-step.
With the linearized space robot dynamics we obtain a trajectory, $\vec{x}_{sp}^*$, that, with the maximal disturbance-robustness degree $\alpha$, satisfies $\phi_{sp}$ with maximal spatial robustness $\rho_{\phi_{sp}}(t,\vec{x})$.
The cost term $J(\vec{u}_{sp}) = \sum_{i,j} |u_i^j|$ approximates total propellant consumption which encourages efficiency\footnote{This comes from the fact that the thrust from thruster $i$ is given by $\vec{F}_i = u_{\mathrm{dc},i}\,\dot m_i \vec{v}_{e,i}$, where $u_{\mathrm{dc},i} \in [0,1]$ is the duty cycle, $\dot m_i$ the propellant mass flow, and $\vec v_{e,i}$ the exhaust velocity. Thus, total propellant consumption is proportional to $\sum_i u_{\mathrm{dc},i}$.} and is of particular interest in space applications where fuel is limited.
The coefficients $c_1$ and $c_2$ are defined such that $c_1 \gg c_2$, and $c_1 \gg \alpha$ to ensure that satisfying the specification with maximal robustness is the primary objective of the optimization problem.
The constraint \eqref{con} ensures the robot respects the effective control bounds. Given that $\mathcal{U}_{spL}$ is a convex polytope in the form $\{u:Hu\leq b\}$ and assuming $\mathcal{D}$ is a zero-centered hyperrectangle, this constraint can be embedded via $\{u \mid Hu \leq \check{b}\}$ with $\check{b}_i = b_i - \alpha \sum(|\boldsymbol{K}^{-T} H_{i,:}|\overline{\mathcal{D}})$ which is linear in its decision variable $\alpha$.% as $K$, $\mathcal{U}$ and $\mathcal{D}$ are known.

\subsection{Underwater Plan Synthesis} %  (prob 3)
\label{ssec:underwater_plan}
% As mentioned, our methodology is to derive a plan $\vec{x}_{uw}^*$ to execute on the underwater platform with dynamics model $\mathcal{M}_{uw}$ that has an equal degree of difficulty as the planned trajectory $\vec{x}_{sp}^*$.
% This is challenging due to the fact that the underwater platform has different dynamics due to both different physical design of the vehicle and different operating environment. 
To achieve a plan, $\vec{x}_{uw}$ with an equal degree of difficult to $\vec{x}_{sp}$, we adjust the mission duration of $\vec{x}_{sp}^*$ such that $\vec{x}_{uw}^*$ has a matching disturbance robustness degree $\alpha^*$ when executed on $\mathcal{M}_{uw}$. 
We adjust the time scaling to find a minimally deviating plan by solving% the following optimization problem:

\begin{align}
\label{eq:problem-mapping}
    \begin{split}
        \min_{\vec{x}_{uw}, \vec{u}_{uw}, \Delta t} & \Delta t, \\
        % \text{s.t.} \quad & \vec{x}_{uw}^{k+1} = \vec{x}_{uw}^{k} + \Delta t(\mathcal{M}_{uw}(\vec{x}_{uw}^{k},\vec{u}_{uw}^{k})),  \\
        \text{s.t.} \quad & \vec{x}_{uw}^{k+1} = \vec{x}_{uw}^k + \mathcal{M}_{uw}(\vec{x}_{uw}^k,\vec{u}_{uw}^k)\Delta t \\
        \quad & (\pos_{uw}^k,\quat_{uw}^k) = (\pos_{sp}^k,\quat_{sp}^k), \\
        \quad & \vec{u}_{uw} \in \mathcal{U}_{uw} \ominus \alpha^* \boldsymbol{K}\mathcal{D}_{uw},
    \end{split}
\end{align}
% where $\text{Euler}(\mathcal{M}_i,\Delta t)$ denotes Euler integration with a $\Delta t$ time step. 
where $(\pos^k_{sp},\quat^k_{sp})$ and $\alpha^*$ are solutions of \cref{eq:problem-def}. 
The resulting trajectory satisfies $\phi_{uw}$, which is a time-scaled version of $\phi_{sp}$ (scaling all intervals in $\phi_{uw}$ with $\frac{\Delta t^\ast}{\Delta t}$).
% Note that as predicates are defined only over $\left[\pos, \quat\right]^T$, the solution of \cref{eq:problem-mapping} is guaranteed to satisfy $\phi_{sp}$ with temporal deviation, governed by $\Delta t^\ast$.

This plan has an equal disturbance robustness degree which means that regardless of whether $\mathcal{M}_{uw}$ is a \emph{faster} or \emph{slower} platform, a motion plan is available for it which can be used to verify operation in space upon observation of these disturbances.

\section{CONTROL}
\label{sec:control}
The planned trajectories include expected bounds on external disturbances which an Extended Kalman Filter (EKF) estimates online for feed-forward control and analysis. 
For underwater validation, we use an MPC with a feedback equivalence law that enforces closed-loop space-robot dynamics.
The MPC for the underwater validation is then defined with respect to the space dynamics, incorporating space-EKF disturbance estimates and tightened actuator limits accounting for the closed-loop model compensation.
Underwater validation then provides practical evidence of feasibility on the space robot, provided that estimated disturbances remain within the assumed bounds.

% The planned trajectories are accompanied with expected bounds on the effect of external disturbances or unmodeled dynamics.
% In our control strategy, we need to estimate/quantify these effects which we do with an Extended Kalman Filter (EKF) formulation. If, after execution, it is observed that these effects lay within the bounds, it would imply sucessfull operation in the space environment under the expected bounds of the offline planner as well. 

% \JV{Note that if we jointly estimate these uncertainties, we might not have a unique solution.}
\begin{figure*}[htb!]
    \centering
    \includegraphics[width=0.98\textwidth]{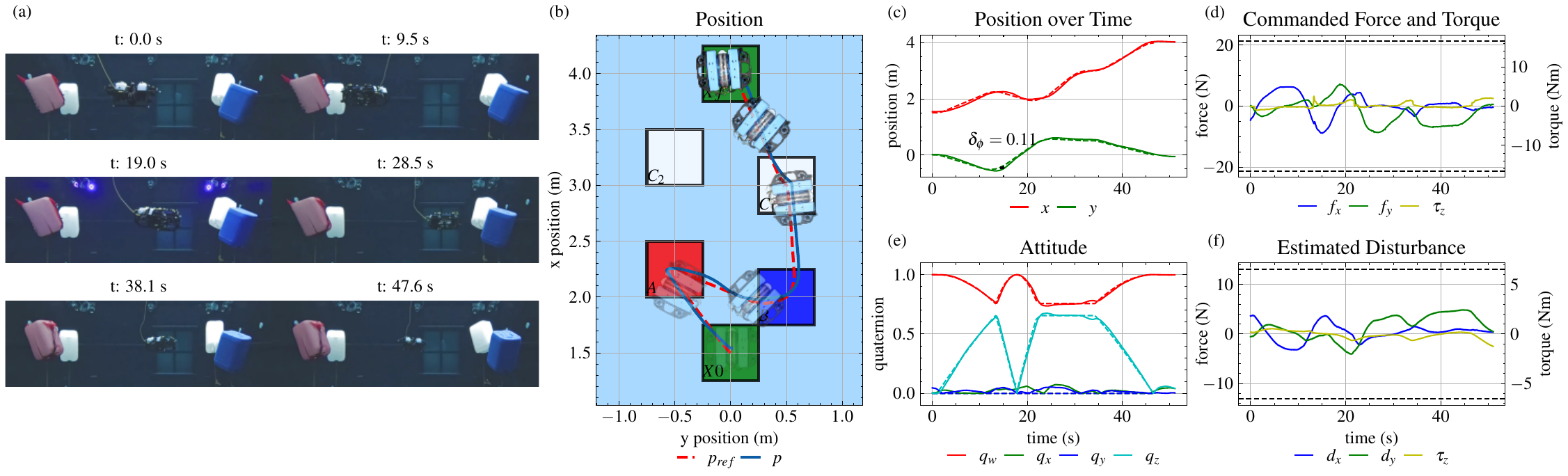}
    \vspace{-3mm}
    % \caption{Hardware results of the Feedback-Equivalent space robot MPC on the BlueROV. The horizontal dotted lines in the Thrust Setpoint indicate $\mathcal{U}$ and in Estimated Disturbance $\alpha^*\mathcal{D}$, indicating that the disturbances lay within the permissible bounds. The maximal spatial deviation is $\delta_{\phi_{uw}}=0.11 \leq \rho_{\phi_{uw}}$.}
    % \caption{Hardware results of the Feedback-Equivalent space-robot MPC of \cref{eq:mpc_uw} on the BlueROV. Dotted traces denote reference trajectories. Horizontal dotted lines mark the input limits $\mathcal{U}_{uw}$ (Commanded Force/Torque) and the admissible disturbance bounds $\alpha^*\mathcal{D}_{uw}$ (Estimated Disturbance). The maximum spatial deviation is $\delta_{\phi_{uw}}=0.11 \leq \rho_{\phi_{uw}}$ which means the specification is satisfied. The observed disturbances being bounded by the disturbance robustness degree with a satisfied specification provides experimental evidence of operation in the space environment, which is shown in \cref{fig:2D-atmos}.}
    \caption{Hardware results of the Feedback-Equivalent underwater-robot MPC of \cref{eq:mpc_uw} on the BlueROV. (a): time-stamped images from the experiment, (b) planned and executed trajectory, (c) position and the maximal spatial deviation $\delta_{\phi_{uw}}=0.11 \leq \rho_{\phi_{uw}} = 0.25$ (indicating satisfaction of $\phi_{uw}$), (d) commanded force and torque with its actuation limits in dotted lines, (e) attitude, (f) estimated force and torque disturbance $\hat{\vec{d}}_{uw}$ with the admissible disturbance bounds $\alpha^*\mathcal{D}_{uw}$. The observed disturbances again being bounded by the disturbance robustness degree $\alpha^*\mathcal{D}_{uw}$ with a satisfied specification provides experimental evidence of operation in the space environment, shown in \cref{fig:2D-atmos}.}
    \label{fig:2D-bluerov}
    \vspace{-3mm}
\end{figure*}
\begin{figure*}[htb!]
    \centering
    \includegraphics[width=0.98\textwidth]{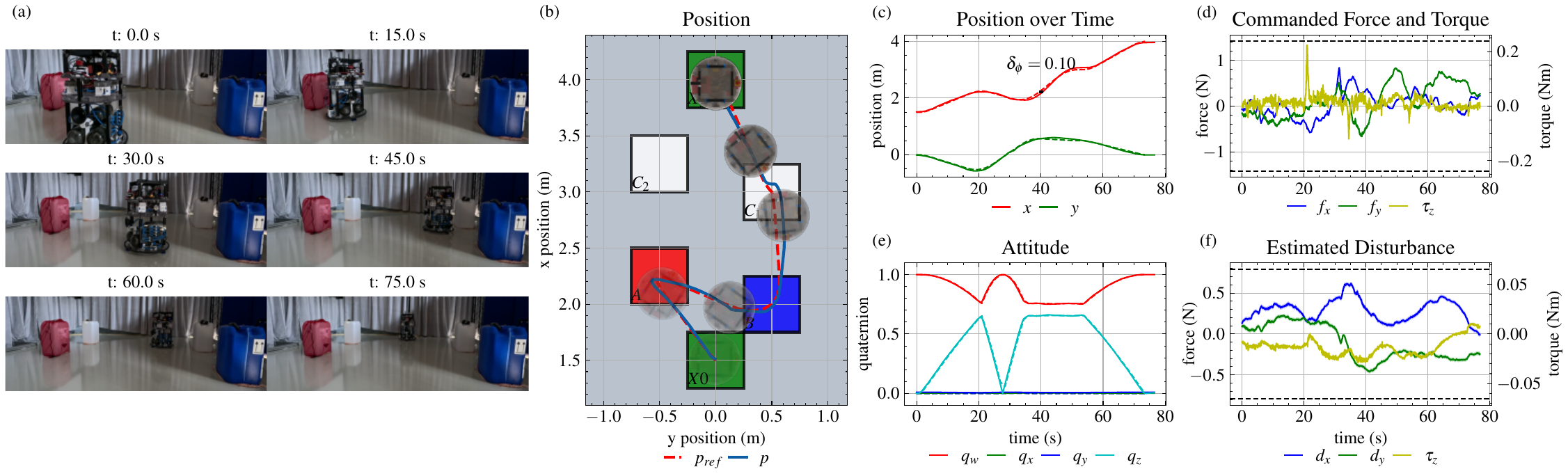}
    \vspace{-3mm}
    % \caption{Hardware results of the space MPC on the space robot. The horizontal dotted lines in the Thrust Setpoint indicate $\mathcal{U}$ and in Estimated Disturbance $\alpha^*\mathcal{D}$, indicating that the disturbances lay within the permissible bounds. The maximal spatial deviation is $\delta_{\phi_{sp}}=0.10 \leq \rho_{\phi_{sp}}$.}
    % \caption{Hardware results of the space-robot MPC of \cref{eq:mpc_sp} on the space robot. Dotted traces denote reference trajectories. Horizontal dotted lines mark the input limits $\mathcal{U}_{sp}$ (Thrust Setpoint) and the admissible disturbance bounds $\alpha^*\mathcal{D}_{sp}$ (Estimated Disturbance). The maximum spatial deviation is $\delta_{\phi_{sp}}=0.10 \leq \rho_{\phi_{sp}}$. The experimental evidence from \cref{fig:2D-bluerov} hence successfully predicts operation in the space environment, based on the maximal disturbance bound from the planner.}
    \caption{Simulation results of the space-robot MPC of \cref{eq:mpc_sp} on ATMOS. 
    (a): time-stamped images from the experiment, (b) planned and executed trajectory in the 2D plane, (c) position and the maximal spatial deviation $\delta_{\phi_{sp}}=0.10 \leq \rho_{\phi_{sp}}=0.25$ (indicating satisfaction of $\phi_{sp}$), (d) commanded force and torque with its actuation limits in dotted lines, (e) attitude, (f) estimated force and torque disturbances $\hat{d}_{sp}$ with the admissible disturbance bounds $\alpha^*\mathcal{D}_{sp}$. 
    % The maximum spatial deviation is $\delta_{\phi_{sp}}=0.10 \leq \rho_{\phi_{sp}}$. 
    The experimental evidence of the underwater validation experiment is translated to the space platform as long as the estimated force and torque disturbances acting on the CubeSat are within the admissible disturbance bounds $\alpha^*\mathcal{D}_{sp}$ which is shown in (f).}
    \label{fig:2D-atmos}
    \vspace{-3mm}
\end{figure*}

\subsection{Disturbance Estimation}
The EKF uses $\mathcal{M}_{\circ}$ with a reduced state augmented with the disturbance force and torque
$\vec{x}^k_{\mathrm{EKF}}=\big[\vec{v}^{k},\;\vec{\omega}^{k},\;\vec{F}_d^{k},\;\vec{\tau}_d^{k}\big]$,
and measurements $\vec{z}^k=\big[\vec{v}^{k},\;\vec{\omega}^{k}\big]$.
Actuator wrenches are assumed equal to commanded values.
% Since actuator forces/torques are not directly measured, we assume we apply the commanded ones.
Disturbances $\vec{d} =[\vec{F}_d,\vec{\tau}_d]$ are modeled as constant over the MPC horizon and as random-walk states in the EKF with zero-mean Gaussian process and measurement noise.
% Process and measurement noises are zero-mean Gaussian, with the process noise including random-walk terms for $\vec{F}_d$ and $\vec{\tau}_d$.
The disturbance estimates are fed forward to the MPC via \(\vec{u}_{\mathrm{ref}}^k=-\hat{\vec{d}}, \forall k \in [0;N-1]\) so the input cost is minimized when the control cancels the estimated disturbance.
See~\cite{roque2025towards} for details.

\subsection{Space robot MPC}
The MPC for the space robot is formulated as
\begin{align}
\label{eq:mpc_sp}
    \begin{split}
        \min_{\vec{x}^{0:N}, \vec{u}^{0:N}, } & \sum_{i=0}^{N-1}||\vec{x}_{sp}^k - \vec{x}^k||_Q + \sum_{i=0}^{N-1}||\vec{u}_{ref}^k - \vec{u}^k||_R + \\ &\quad\quad||\vec{x}_{sp}^N-\vec{x}^N||_P, \\
        \text{s.t.} \quad & \vec{x}^{k+1} = \text{RK4}(\mathcal{M}_{sp}(\vec{x}^k,\vec{u}^k,\hat{\vec{d}})),  \\
        & \vec{x}^k \in \mathcal{W}_{\text{free}} \\
        & \vec{u}^k \in \mathcal{U}_{sp} %u^k \in \alpha \bar{\mathcal{U}}_{sp},
    \end{split}
\end{align}
where $Q\succeq0$ and $R \succ 0$ are stage cost weight matrices and $P \succeq 0$ is the terminal weight matrix, $\text{RK4}$ denotes the Runge-Kutta 4 integration scheme~\cite{schwartz1996theory}, and $\mathcal{W}_{\text{free}}$ denotes the workspace. Obstacle avoidance is captured in $\phi_{\circ}$, and $\vec{x}_{sp}$ is obtained from the planner.
% To estimate external force and torque disturbances, we employ a parameter-estimating Extended Kalman Filter (EKF) using $\mathcal{M}_{sp}$.

% It should be noted that in accordance with Assumption~\ref{ass:observability}, we assume in our guarantees that the EKF converges instantaneously.
% \JV{At the end we have a state and control trace and a disturbance mean and covariance estimate. Perhaps we can use this for a probabilistic guarantee.}

\subsection{Dynamical Equivalency}
% While underwater environments are often considered as close analogues to space, there are significant and fundamental differences as shown in Sec.~\ref{sec:ffmodels}. 
% While underwater robots can be considered close analogues to space robots, it is in fact only neutrally buoyant, undisturbed, static objects that (locally) behave equivalent to systems in space.

We apply feedback equivalence control to $\mathcal{M}_{uw}$ to behave as $\mathcal{M}_{sp}$ in closed-loop. 
This compensates for effects such as hydrodynamics and may even introduce orbital dynamics to $\mathcal{M}_{uw}$. 
The transformation is defined as
% The result is that during the experimental validation, we are able to run a controller that considers the dynamics of the space robot with tightened actuator constraints for the compensating behavior.
% As $\mathcal{M}_{sp}$ and $\mathcal{M}_{uw}$ are in control-affine form we define
\begin{equation}
\label{eq:fbl}
    \vec{u}_{sp\rightarrow uw}(\vec{x},\vec{u}) = g_{uw}(\vec{x})^{\dagger}(f_{sp}(\vec{x}) + g_{sp}(\vec{x})\vec{u} - f_{uw}(\vec{x}))
\end{equation}
% \begin{equation}
% \label{eq:fbl}
% \begin{aligned}
%     \vec{u}_{sp\rightarrow uw}(\vec{x},\vec{u}_{sp}) 
%     &= g_{uw}(\vec{x})^{\dagger}\big(f_{sp}(\vec{x})  \\
%     &\quad + g_{sp}(\vec{x})\vec{u}_{sp} - f_{uw}(\vec{x})\big)
% \end{aligned}
% \end{equation}
where $g_{uw}^{\dagger}$ denotes the pseudo-inverse.
% \CRS{I thought we decided to only apply feedback equivalence to the dynamics not the kinematics?}
This controller ensures $\mathcal{M}_{uw}$ internally behaves like $\mathcal{M}_{sp}$: the output of $\vec{u}_{sp\rightarrow uw}$ is the input that needs to be given to the underwater robot such that it behaves as if $\vec{u}$ is applied to the space robot.

\subsection{Underwater robot MPC}
The MPC for the underwater robot is defined as
\begin{align}
\label{eq:mpc_uw}
    \begin{split}
        \min_{\vec{x}^{0:N}, \vec{u}^{0:N}, } & \sum_{i=0}^{N-1}||\vec{x}_{uw}^k - \vec{x}^k||_Q + \sum_{i=0}^{N-1}||\vec{u}_{uw}^k - \vec{u}^k||_R + \\&\quad\quad||\vec{x}_{uw}^N-\vec{x}^N||_P, \\
        \text{s.t.} \quad & \vec{x}^{k+1} = \text{RK4}(\mathcal{M}_{sp}(\vec{x}^k,\vec{u}^k,\hat{\vec{d}})),  \\
        & \vec{x}^k \in \mathcal{W}_{\text{free}} \\
        & \vec{u}_{sp\rightarrow uw}(\vec{x}^k,\vec{u}^k) \in \mathcal{U}_{uw}
        % & \vec{u}_{sp\rightarrow uw}(\vec{x}^k,\underline{\mathcal{U}}_{uw}) \leq \vec{u}^k \leq \vec{u}_{sp\rightarrow uw}(\vec{x}^k,\bar{\mathcal{U}}_{uw})
    \end{split}
\end{align}
% \begin{align}
% \label{eq:mpc_uw}
%     \begin{split}
%         \min_{x^{0:N}, u^{0:N}, } & \sum_{i=0}^{N}||x_{ref}^k - x^k||_Q + \sum_{i=0}^{N-1}||u_{ref}^k - u^k||_R, \\
%         \text{s.t.} \quad & x^{k+1} = f_{sp}(x^k) + g_{sp}(x^k)u^k + C\hat{d},  \\
%         \quad & u^k \in u_{uw\rightarrow ff}(x^k,\mathcal{U}_{uw}),
%     \end{split}
% \end{align}
which utilizes the dynamics of the space robot with the tightened constraints from the feedback-equivalence control law in \cref{eq:fbl}, similar to~\cite{simon2013nonlinear}. 
% The disturbance estimation may then be achieved by an EKF using $\mathcal{M}_{sp}$.
The final control input to the underwater robot is then obtained as $\vec{u}_{uw} = \vec{u}_{sp\rightarrow uw}(\vec{x}^0,\vec{u}^0)$. 
While \cref{eq:mpc_uw} utilizes a disturbance estimator using $\mathcal{M}_{sp}$, the requirement of $\vec{d} \in \mathcal{D}_{uw}$ is obtained from observing a parallel disturbance estimator using $\mathcal{M}_{uw}$.
% Then, upon observation that $\hat{\vec{d}} \in \mathcal{D}_{uw}$ and $\vec{x}_{uw} \models \phi_{uw}$, it is implied that $\vec{x}_{sp} \models \phi_{sp}$ using the controller in \cref{eq:mpc_sp} under Assumptions~\ref{ass:column_space}, and ~\ref{ass:supset}.
% \JV{A note pointing back to the proof and the affect that the MPC and EKF have on that!}

% \JV{
% Comments during meeting:
% if we assume state is fully known, only then could we get ground truth + ground truth cov of the disturbance estimate (solving the least-squares problem). EKF with state + disturbance is a reasonable approach.

% Tighten the constraints with the covariance.

% Could potentially include the EKF in the MPC and minimize the trace of the covariance estimate.
% }

\section{VALIDATION}\label{sec:validation}
% The goal of our experimental validation is to validate to what extent we can utilize underwater robots as testing facilities for space robots for high-level planning and control.
% To recap, the goal of this experiment is to rigorously validate a space mission using an underwater robot.

\subsection{Experimental Setup}
% In this section, we present the experimental setups for the space and underwater environments. 
% Both robots utilize an identical software stack where the state estimation is performed by fusing onboard IMU measurements with offboard motion-capture data. Then high-level force and torque commands from solving \cref{eq:mpc_sp} or \cref{eq:mpc_uw} with ACADOS~\cite{Verschueren2021} are obtained from an offside computer and sent to a Pixhawk 6X–mini running PX4~\cite{}. This microcontroller computes low–level control by allocating the commanded forces and torques to PWM signals to either the thrusters or propellers.
% Ground-truth state estimation is used in simulation, while motion capture is assumed to provide ground-truth state estimation on hardware.

% We utilize the tracking for state estimation to mitigate the fact that state estimation in space and underwater is treated differently with its own set of challenges, which are outside the scope of this paper.

% The entire software stack is identical for both robots with 
% facilities have a Qualisys motion capture to track the respective robot. 

Experiments use an underwater robot as the validation of a space-analog platform, all running an identical software stack: a Pixhawk 6X–mini microcontroller, running PX4~\cite{meier_px4_2015}, performs sensor fusion of onboard IMU and external motion-capture data, and computes low-level control by allocating commanded forces and torques to PWM signals for the thrusters. Offboard computers solve \cref{eq:mpc_sp} or \cref{eq:mpc_uw} with ACADOS~\cite{Verschueren2021}, providing wrench commands that are then transmitted to the microcontroller.
For all our experiments the timestep and horizon length are identical but weight matrices $Q$, $R$, and $P$ are altered between \cref{eq:mpc_sp} and \cref{eq:mpc_uw} to improve stability of the controllers.

\subsubsection*{Underwater Setup}
For underwater validation, we use a BlueRobotics BlueROV2 Heavy in the KTH Marinarium testing facilities~\cite{torroba2026marinariumnewarenabring}. The dynamics model is parameterized according to~\cite{von2022open}, with actuator limits reduced to avoid cavitation: $\overline{\mathcal{U}}_{uw}=-\underline{\mathcal{U}}_{uw}=[21,\,21,\,30,\,20,\,11,\,17]$, and disturbance bounds $\overline{\mathcal{D}}_{uw}=-\underline{\mathcal{D}}_{uw}=[10,\,10,\,10,\,5,\,5,\,5]$.

\subsubsection*{Space Setup}
Validation involves (i) The ATMOS air-bearing platform~\cite{roque2025towards} for 3-DoF frictionless 2D dynamics with compressed-air thrusters and (ii) A Basilisk simulation~\cite{kenneally2020basilisk} with a CubeSat parameterized to match the space robot's inertia and actuator configuration, placed in a 700\,km circular equatorial orbit and propagated in the rotating Hill frame. 
A bridge interface exposes Basilisk's internal message system and provides a PX4-compatible communication layer~\cite{krantz2025bridging}. 
Both environments share actuator limits
$\overline{\mathcal{U}}_{sp}=-\underline{\mathcal{U}}_{sp}=[1.42,\,1.42,\,1.42,\,0.24,\,0.24,\,0.24]$, which is below the hardware maximum to allow simultaneous force and torque saturation, and
$\overline{\mathcal{D}}_{sp}=-\underline{\mathcal{D}}_{sp}=[0.64,\,0.64,\,0.64,\,0.01,\,0.01,\,0.01]$,
where forces are assumed from floor imperfections, and torques are assumed to be small.

\subsection{Results}
We perform validation in two inspection-style experiments: 1)  a 2D mission, comparing the BlueROV with the physical space robot, and 2) a 3D mission, comparing the BlueROV with the simulated CubeSat.

% We perform validation in two experiments: 1) a planar inspection mission on the BlueROV to validate execution of a planar experimental robotic platform, 2) a 6-DoF inspection mission on the BlueROV to validate execution of a simulated CubeSat mission.

\subsubsection{Planar Specification}
Consider an intra-vehicular space robot (e.g., the Astrobee~\cite{fluckiger_astrobee_2018} in the ISS) tasked with visually inspecting certain areas of interest within certain time intervals.
The specification for the space robot is
\begin{multline}
\label{eq:stl_exp1}
    \phi_{sp} = \diamondsuit_{[20,25]}(\vec{\eta}\in A) \land \diamondsuit_{[0,37]}(\vec{\eta} \in B) \\\land (\Box_{[50,55]}(\vec{\eta}\in C_1) \lor \Box_{[50,55]}(\vec{\eta} \in C_2))
\end{multline}
with $\vec{\eta}(0)\in X_0$ and $\vec{\eta}(75) \in X_f$. The regions of interest $X_0$, $X_f$, $A$, $B$, $C_1$, and $C_2$ (including desired orientations) are defined in Tab.~\ref{tab:spec_details} and shown in \cref{fig:2D-atmos}. This specification captures that both the red and blue region should be visited within a certain interval while one of the white regions has to be investigated for at least $5$ seconds.
% Due to the linearity requirements of the predicates in the MILP planner, we $k \in \{A,B\}$ as sets containing a position and an orientation such that the area of interest is in vision (e.g. $A=\{\pos_x \in [-1,1],\pos_y\in[-0.2,0.2],\quat_{\textrm{yaw}}\in[-\frac{\pi}{8},\frac{\pi}{8}]\}$ as opposed to nonlinear vision-cone constraints).

We first obtain the nominal motion plan for the space robot by solving \cref{eq:problem-def}. The result is shown in \cref{fig:2D-atmos} with the disturbance-robustness degree of $\alpha^*=1.31$, indicating that $\exists \vec{u} \in \mathcal{U}, s.t. \forall \vec{d} \in 1.31\cdot\mathcal{D}, \vec{x}_{sp}\models \phi_{sp}$.
We then obtain the scaled motion plan for the underwater robot according to Eq.~\eqref{eq:problem-mapping} shortening the mission duration $t_f$ from $75$ to $65.5$ seconds (a $1.15\times$ speedup as the control limits of the BlueROV model are higher than the space robot), making the underwater plan have an identical disturbance-robustness degree of $\alpha^*=1.31$, shown in \cref{fig:2D-bluerov}.
During execution on the space robot, we solve \cref{eq:mpc_sp} with the estimated disturbance from the space robot EKF. As this control strategy is to be validated on the different underwater robot, we solve the feedback-equivalent form in \cref{eq:mpc_uw}, which considers the space robot model and a space robot EKF.
The experimental results on the BlueROV in \cref{fig:2D-bluerov}. \cref{fig:2D-bluerov}(f) indicate that the estimated disturbances on the BlueROV are within the $\alpha^*\mathcal{D}_{uw}$ bounds. \cref{fig:2D-bluerov}d shows the control inputs and the $\overline{\mathcal{U}}_{uw}=\mathcal{U}_{uw} \ominus \alpha^*\mathcal{D}_{uw}$ bounds.
The results, together with the assumptions from the planner, imply operation on the space robot as $\hat{\vec{d}}_{sp} \in \mathcal{D}_{sp}^*$, as shown in \cref{fig:2D-atmos}.
% , where the space robot successfully completes its mission using the control strategy in \cref{eq:mpc_sp}.

\begin{figure*}[t]
    \centering
    \includegraphics[width=0.98\textwidth]{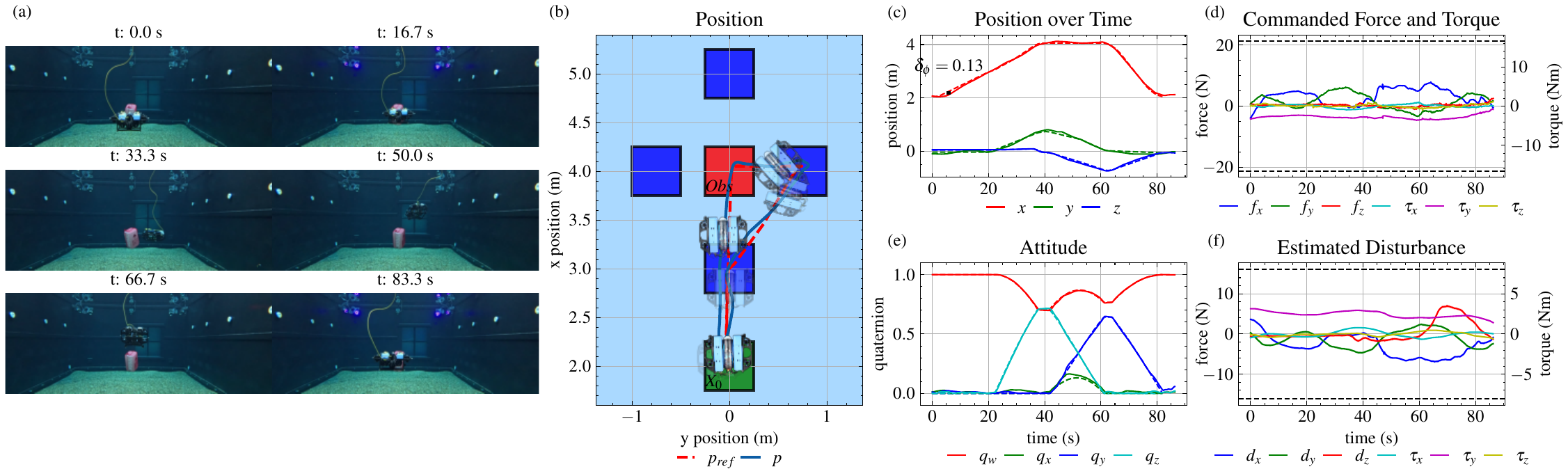}
    % \caption{Hardware results of the Feedback-Equivalent CubeSat MPC on the BlueROV platform. Note that the horizontal dotted lines in the Thrust Setpoint indicate $\mathcal{U}$ and in Estimated Disturbance $\alpha^*\mathcal{D}$, indicating that the disturbances lay within the permissible bounds. The maximal spatial deviation is $\delta_{\phi_{uw}}=0.13 \leq \rho_{\phi_{uw}}$.}
    \vspace{-3mm}
    \caption{Hardware results of the Feedback-Equivalent CubeSat MPC of \cref{eq:mpc_uw} on the BlueROV. (a): time-stamped images from the experiment, (b) planned and executed trajectory in the 2D plane, (c) position and the maximal spatial deviation $\delta_{\phi_{uw}}=0.13 \leq \rho_{\phi_{uw}}=0.19$ (indicating satisfaction of $\phi_{uw}$), (d) commanded force and torque with its actuation limits in dotted lines, (e) attitude, (f) estimated force and torque disturbances $\hat{d}_{uw}$ with the admissible disturbance bounds $\alpha^*\mathcal{D}_{uw}$. The observed disturbances again being bounded by the disturbance robustness degree $\alpha^*\mathcal{D}_{uw}$ with a satisfied specification provides experimental evidence of operation in the space environment, shown in \cref{fig:3D-atmos}.}
    \label{fig:3D-bluerov}
    \vspace{-3mm}
\end{figure*}
\begin{figure*}[t]
    \centering
    \includegraphics[width=0.98\textwidth]{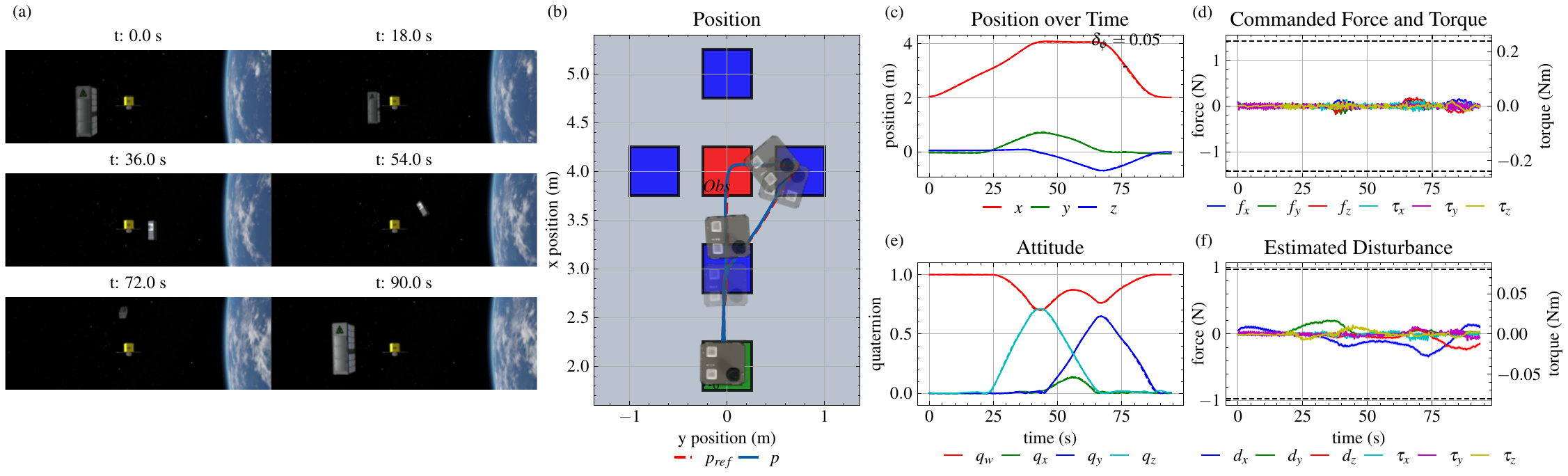}
    % \caption{Simulation results of the CubeSat MPC on the CubeSat platform. The horizontal dotted lines in the Thrust Setpoint indicate $\mathcal{U}$ and in Estimated Disturbance $\alpha^*\mathcal{D}$, indicating that the disturbances lay within the permissible bounds. The maximal spatial deviation is $\delta_{\phi_{sp}}=0.05 \leq \rho_{\phi_{sp}}$.}
    \vspace{-3mm}
    \caption{Simulation results of the CubeSat MPC of \cref{eq:mpc_sp} on the CubeSat. (a): time-stamped images from the experiment, (b) planned and executed trajectory in the 2D plane, (c) position and the maximal spatial deviation $\delta_{\phi_{sp}}=0.05 \leq \rho_{\phi_{sp}}=0.19$ (indicating satisfaction of $\phi_{sp}$), (d) commanded force and torque with its actuation limits in dotted lines, (e) attitude, (f) estimated force and torque disturbances $\hat{d}_{sp}$ with the admissible disturbance bounds $\alpha^*\mathcal{D}_{sp}$. 
    % The maximum spatial deviation is $\delta_{\phi_{sp}}=0.05 \leq \rho_{\phi_{sp}}$. 
    The experimental evidence of the underwater validation experiment is translated to the space platform as long as the estimated force and torque disturbances acting on the CubeSat are within the admissible disturbance bounds $\alpha^*\mathcal{D}_{sp}$, shown in (f).}
    \label{fig:3D-atmos}
    \vspace{-3mm}
\end{figure*}

\subsubsection{Complex Inspection Task}
Consider a task where an extra-vehicular CubeSat needs to inspect the sides of an unknown satellite. 
The specification captures that each axis of the satellite is to be observed, e.g., the $x$ axis of the spacecraft has to be observed from the front or the left side. 
The specification for the space robot is
\begin{equation}
\label{eq:stl_exp2}
    \phi_{sp} = \bigwedge_{k={x,y,z}}\diamondsuit_{[\underline{t}_k,\overline{t}_k]}(\vec{\eta} \in \underline{A}_k \lor \vec{\eta} \in \overline{A}_k) \land \Box_{[0,90]}(\vec{\eta} \notin \textrm{Obj})
\end{equation}
with $\vec{\eta}(0) \in X_0$ and $\vec{\eta}(90) \in X_0$. The regions of interest $\underline{A}_k$ and $\overline{A}_k$ (including required rotation) and their time intervals are specified in Tab.~\ref{tab:spec_details}.
The nominal motion plan has a disturbance-robustness degree of $\alpha^*=1.61$. The scaled motion plan for the underwater robot reduces the mission duration $t_f$ from $90$ to $80.4$ seconds (a speed-up of $1.11\times$, less significant due to the torque requirements on the BlueROV model).

The results on the BlueROV in \cref{fig:3D-bluerov} show that the torque disturbance is particularly large around $y$ (likely due to a model mismatch of the center of buoyancy and center of mass) yet still within the permissible disturbance bounds of $\alpha^*\mathcal{D}_{uw}$. The satisfaction $\phi_{uw}$ with disturbances within this scaled set shows experimental evidence that the planning and control architecture will operate successfully in space, under the same disturbance robustness bounds.
Using the EKF, we are able to quantify this modeling error (plus any additional disturbances from, for example, the tether). 
% Upon observation that the estimated disturbances do not exceeds the $\alpha^*\mathcal{D}$ bounds
In the validation of the CubeSat mission in Basilisk in \cref{fig:3D-atmos}, the disturbances are significantly below $\alpha^*\mathcal{D}$. 
% The underwater environment however incorporates hardware and software elements of the space platform which can be tested.
% If this model is significantly off, the estimated disturbances will not lay within the planned bounds. 
% As such, from \cref{fig:3D-bluerov} we cannot conclude operation on the CubeSat. 

\subsection{Discussion}
The results demonstrate that space planning and control can be validated on dynamically equivalent underwater platforms, by quantifying permissible disturbances via the disturbance robustness degree while the estimator verifies online whether this assumption holds.
% the promise of our approach to validating space robotics in underwater environments. We enable verification of planning and control algorithms on dynamically equivalent yet different platforms as long as estimated disturbances are within planned bounds.
% As the planner provides the minimal robustness assumption (by maximizing the disturbance set), the estimator online verifies whether these assumptions are met on deployment.

There are some limitations however. First, the guarantees hold with the assumptions in Sec.~\ref{sec:planning}, of which instantaneously detecting and counteracting the disturbance are the most significant.
Additionally, space missions can only be validated with underwater robots if the underwater model is sufficiently accurate, such that its errors can be represented as additive disturbances within the maximal disturbance robustness degree. However, the permissible level of mismatch is not identified.
Lastly, uncertainty and different characteristics of estimation pipelines are currently not considered as ground-truth state estimation is assumed.
\section{CONCLUSIONS}
% \textcolor{red}{TODO}
We presented an experimental framework for the validation of space robotics in underwater environments. 
Using disturbance robustness as an optimization metric, we generate motion plans that satisfy a high-level specification while being of equal disturbance robustness degree for the space robot and the underwater analogue. 
This ensures that, regardless of the dynamical and environmental differences, a motion plan for the validation platform can be obtained.
Estimating the overall disturbances to the platform in the underwater analogue platform then provides practical evidence that planning and control methods are able to operate in the inaccessible space environment as long as these disturbances are bounded by the maximal disturbance set.

Our theoretical contribution relies on instantaneous disturbance estimation and rejection. Future work will relax these assumptions. 
Additionally, we will extend our method with time-varying disturbance robustness metrics, complex and multi-robot scenarios such as docking and transportation, and enable human-in-the-loop validation of space missions.

\setlength{\tabcolsep}{5pt}
\begin{table}[]
\caption{Details on the STL specifications from \cref{eq:stl_exp1} and \cref{eq:stl_exp2}.}
\label{tab:spec_details}
\begin{tabular}{lllll}
\toprule
                  & center               & widths            & dimensions$^*$ & interval   \\
\midrule
A                 & $[2.25,-0.5,-\frac{\pi}{2}]$ & $[0.5,0.5,\frac{\pi}{4}]$ & [$x$,$y$,$\psi$] & [20,25]\\ 
B                 & $[2,0.5,-\frac{\pi}{2}]$     & $[0.5,0.5,\frac{\pi}{4}]$ & [$x$,$y$,$\psi$] & [0,37]\\ 
$C_1$             & $[3,0.5,\frac{\pi}{2}]$      & $[0.5,0.5,\frac{\pi}{4}]$ & [$x$,$y$,$\psi$] & [50,55]\\ 
$C_2$             & $[3.25,-0.5,\frac{\pi}{2}]$  & $[0.5,0.5,\frac{\pi}{4}]$ & [$x$,$y$,$\psi$] & [50,55]\\ 
$\underline{A}_x$ & $[3,0,0,0,0]$        & $[0.5,0.5,\frac{\pi}{8},\frac{\pi}{8}]$ & [$x$,$y$,$z$,$\theta$,$\psi$] & [20,25] \\ 
$\overline{A}_x$       & $[5,0,0,0,\pi]$      & $[0.5,0.5,\frac{\pi}{8},\frac{\pi}{8}]$ & [$x$,$y$,$z$,$\theta$,$\psi$] & [20,25] \\ 
$\underline{A}_y$ & $[4,-0.75,0,0,\frac{\pi}{2}]$ & $[0.5,0.5,\frac{\pi}{8},\frac{\pi}{8}]$ & [$x$,$y$,$z$,$\theta$,$\psi$]  & [40,45]\\ 
$\overline{A}_y$       & $[4,0.75,0,0,\frac{\pi}{2}]$  & $[0.5,0.5,\frac{\pi}{8},\frac{\pi}{8}]$ & [$x$,$y$,$z$,$\theta$,$\psi$]  & [40,45]\\ 
% $\underline{A}_z$ & $[4,0,-0.75,0,\frac{\pi}{2},0]$ & $[0.5,0.5,\frac{\pi}{8},\frac{\pi}{8}]$ & [$x$,$y$,$z$,$\theta$,$\psi$]  & [65,70]\\ 
$\overline{A}_z$       & $[4,0,0.75,0,-\frac{\pi}{2},0]$ & $[0.5,0.5,\frac{\pi}{8},\frac{\pi}{8}]$ & [$x$,$y$,$z$,$\theta$,$\psi$]  & [65,70]\\
\bottomrule
\end{tabular}

\medskip
\footnotesize{$^*$Here, $\theta$ and $\psi$ denote pitch and yaw.}
\vspace{-5mm}
\end{table}

\bibliographystyle{IEEEtran}
\bibliography{references}

\end{document}